\documentclass{article}
\usepackage{arxiv}

\usepackage{url,lineno,microtype}
\usepackage[onehalfspacing]{setspace}
\usepackage{textcomp}
\usepackage{graphicx}
\usepackage[numbers]{natbib}
\begin{document}

\title{Deep Learning for Whole Slide Image Analysis: An Overview} 

\author{
  Neofytos Dimitriou\\
  School of Computer Science\\
  University of St Andrews\\
  St Andrews, United Kingdom \\
  \texttt{neofytosd@gmail.com} \\
   \And
  Ognjen Arandjelovi\'c\ \\
  School of Computer Science\\
  University of St Andrews\\
  St Andrews, United Kingdom \\
  \texttt{ognjen.arandjelovic@gmail.com} \\
  \And
  Peter D Caie\\
  School of Medicine\\
  University of St Andrews\\
  St Andrews, United Kingdom \\
  \texttt{pdc5@st-andrews.ac.uk} \\
}



\maketitle

\begin{abstract}
The widespread adoption of whole slide imaging has increased the demand for effective and efficient gigapixel image analysis. Deep learning is at the forefront of computer vision, showcasing significant improvements over previous methodologies on visual understanding. However, whole slide images have billions of pixels and suffer from high morphological heterogeneity as well as from different types of artefacts. Collectively, these impede the conventional use of deep learning. For the clinical translation of deep learning solutions to become a reality, these challenges need to be addressed. In this paper, we review work on the interdisciplinary attempt of training deep neural networks using whole slide images, and highlight the different ideas underlying these methodologies.
\end{abstract}
\tiny
\normalsize

\section{Introduction}

The adoption of digital pathology into the clinic will arguably be one of the most disruptive technologies introduced into the routine working environment of pathologists. Digital pathology has emerged with the digitization of patient tissue samples and in particular the use of digital whole slide images (WSIs). These can be distributed globally for diagnostic, teaching, and research purposes. Validation studies have shown correlation between digital diagnosis and glass based diagnosis~\cite{SneaTsanMeskKima+2015,PantSinaHenrFath+2013}. However, although multiple whole slide scanners are currently available on the market, to date only Philips Ultra-fast scanner has been approved by regulatory bodies for the use in primary diagnosis~\cite{Cacco2017}. 

Recently, NHS Greater Glasgow and Clyde, one of the largest single site pathology services in Europe, has begun proceedings to undergo full digitization. As the adoption of digital pathology becomes wider, automated image analysis of tissue morphology has the potential to further establish itself in pathology and ultimately decrease the workload of pathologists, reduce turnaround times for reporting, and standardise clinical practices. For example, known or novel biomarkers and histopathological features can be automatically quantified  ~\cite{HamiBankWangHutc+2014,HardSchoNekoMeie+2019,JanoMada2016,BrieGavrNearHarr+2019,CaieZhouTurnOnis+2016,BeckSangLeunMari+2011,NearLillGavrUeno+2019, SariGund2019,DimiAranHarrCaie2018}. Furthermore, deep learning techniques can be employed to recognize morphological patterns within the specimen for diagnostic and triaging purposes~\cite{QuasMukhReddMunu+2017,WangKhosGargIrsh+2016,LiuGadeNoroDahl+2017,YueDimiCaieHarr+2019}.

Successful application of deep learning to WSIs has the potential to create new clinical tools that surpass current clinical approaches in terms of accuracy, reproducibility, and objectivity while also providing new insights on various pathologies. However, WSIs are multi-gigabyte images with typical resolutions of $100,000 \times 100,000$ pixels, present high morphological variance, and often contain various types of artefacts. These conditions preclude the direct application of conventional deep learning techniques. Instead, practitioners are faced with two non-trivial challenges. On the one hand, the visual understanding of the images, impeded by the morphological variance, artefacts, and typically small data sets, and, on the other hand, the inability of the current state of the hardware to facilitate learning from images with such high resolution, thereby requiring some form of dimensionality reduction to the images. These two problems are sometimes referred to as the \textit{what} and \textit{where} problems~\cite{AichGhas2018,TellLitjLaakCiom2018}. In this paper, we first discuss important aspects and challenges of WSIs, and then delve deeper into the different approaches to the two aforementioned problems.

\section{Whole slide images}
\subsection{Tissue visualization}
The majority of WSIs are captured using brightfield illumination, such as for slides stained with clinically routine haematoxylin and eosin (H\&E). 
The wider accessibility of H\&E stained WSIs, compared to more bespoke labelling reagents, at present makes this modality more attractive for deep learning applications. H\&E stained tissue is excellent for the characterization of morphology within a tissue sample which corroborates to its long use in clinical practice.

However, H\&E stained slides lack \textit{in situ} molecular data associated with a cell. In contrast, this is possible with protein visualization through immunolabelling. The labelling of multiple cell types and their protein expression can be observed with multiplexed immunofluorescence (IF) which provides valuable information in cancer research and particularly in immunooncology~\cite{BrieGavrNearHarr+2019,NearLillGavrUeno+2019,WongWeiSmitAcs+2019}. Nevertheless, the analysis of IF labelled slides using deep learning techniques is impeded by the limited availability of IF WSI data sets. Potential causes of the data scarcity include the expense of reagents and of access to fluorescence scanners, as well as the enormous IF WSI size which can sometimes exceed 10 gigabytes per image.

\subsection{Data availability}
Unlike in numerous other fields which have adopted supervised deep learning techniques \cite{SchlAran2017,CoopAran2019}, labelled data is more difficult to obtain in digital pathology, thereby challenging the practicability of supervised approaches. Despite wider data publication in the recent years~\cite{SiriDomiRichRed+2019,EhteVetaJoha+2017,BandGeesMansDijk+2019,AresArauKwokChen+2018,QuasMukhReddMunu+2017,VetaHengStatBejn+2019}, much of the published work still employs proprietary WSI data sets~\cite{CampSilvFuch2018}.


\subsection{Image format}
There are currently multiple whole slide scanners from different vendors available on the market with the capacity for both brightfield and fluorescence imaging. Each scanner captures images using different compression types and sizes, illumination, objectives, and resolution and also outputs the images in a different proprietary file format. The lack of a universal image format can delay the curation of large data sets. The field of radiology has overcome this issue with the adoption of DICOM open source file formats allowing large image data sets to be accessed and interrogated~\cite{BennSmitJaroNola+2018,KahnCarrFlynPeck+2007}. Digital pathology is yet to widely adopt a single open source file format although work and discussions are continually progressing towards this end~\cite{LennHerrClunFedo+2018,Clun2019}.

\subsection{Artefacts and colour variability}
\begin{figure}[h!]
\begin{center}
\includegraphics[width=15cm]{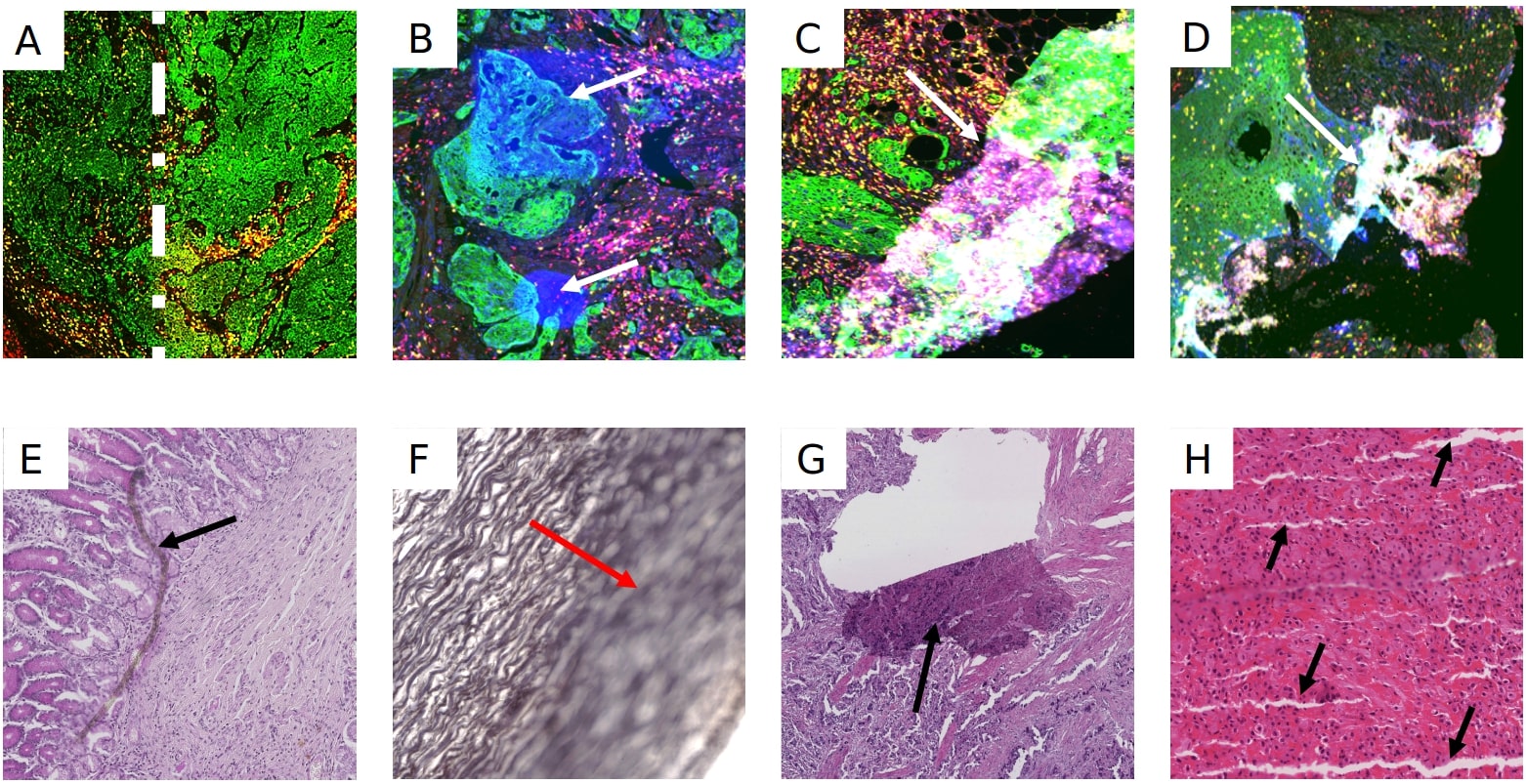}
\end{center}
\caption{Examples of artefact in both fluorescence and brightfield captured images. Images A--D are examples of multiplex IF images containing different types of artefacts. These images were taken from slides labelled with Pan-cytokeratin (green), DAPI (blue), CD3 (yellow), and CD8 (red). (A): Higher intensity of Pan-cytokeratin on the right region than the left as defined by the dotted white line. (B): White arrows point to high intensity regions in the DAPI channel artificially produced during imaging. (C, D): White arrows show tears and folds in the tissue that result in out of focus and fluorescence artefacts. Images E--H contain examples of artefacts from brightfield captured images labelled with H\&E (E, G, and H) or Verhoeff's elastic stain (F). (E): Black arrow highlights foreign object under coverslip. (F): Red arrow highlights out of focus region. (G): black arrow shows tear in tissue. (H): black arrows show cutting artefacts. All images were captured with a $20\times$ objective on a Zeiss Axioscan.z1.}
\label{fig:1}
\end{figure}

To be clinically translatable, deep learning algorithms must work across large patient populations and generalize over image artefacts and colour variability in staining~\cite{CaieSchuOnisMull+2013}.
Artefacts can be introduced throughout the entire sample preparation workflow as well as during the imaging process. These can include ischemia times, fixation times, microtome artefact, staining reagent variability as well as imaging artefacts from uneven illumination, focusing, image tiling and fluorescence deposits and bleed-through. Examples of such artefacts are shown in Figure~\ref{fig:1}. 

Through training, the human brain can become adept at ignoring artefacts and staining variability, and honing in the visual information necessary for an accurate diagnosis. To facilitate an analogous outcome in deep learning models, there are generally two approaches that can be followed. The first involves explicit removal of artefacts (e.g.\ using image filters), as well the normalisation of colour variability~\cite{MageTreaCrelShir+2009}. In contrast, the second approach takes on a less direct strategy, augmenting data with often synthetically generated data which captures a representative variability in artefacts and staining, making their learning an integral part of the training process. Both approaches have been employed with some success to correct the variation from batch effect or from archived clinical samples from different clinics~\cite{BrieCaieGavrSchm+2018} though this finding has not been universal~\cite{LiuGadeNoroDahl+2017}.

\section{Deep Learning}
\subsection{Patch extraction}
\begin{figure}[htbp]
\begin{center}
\includegraphics[width=15cm]{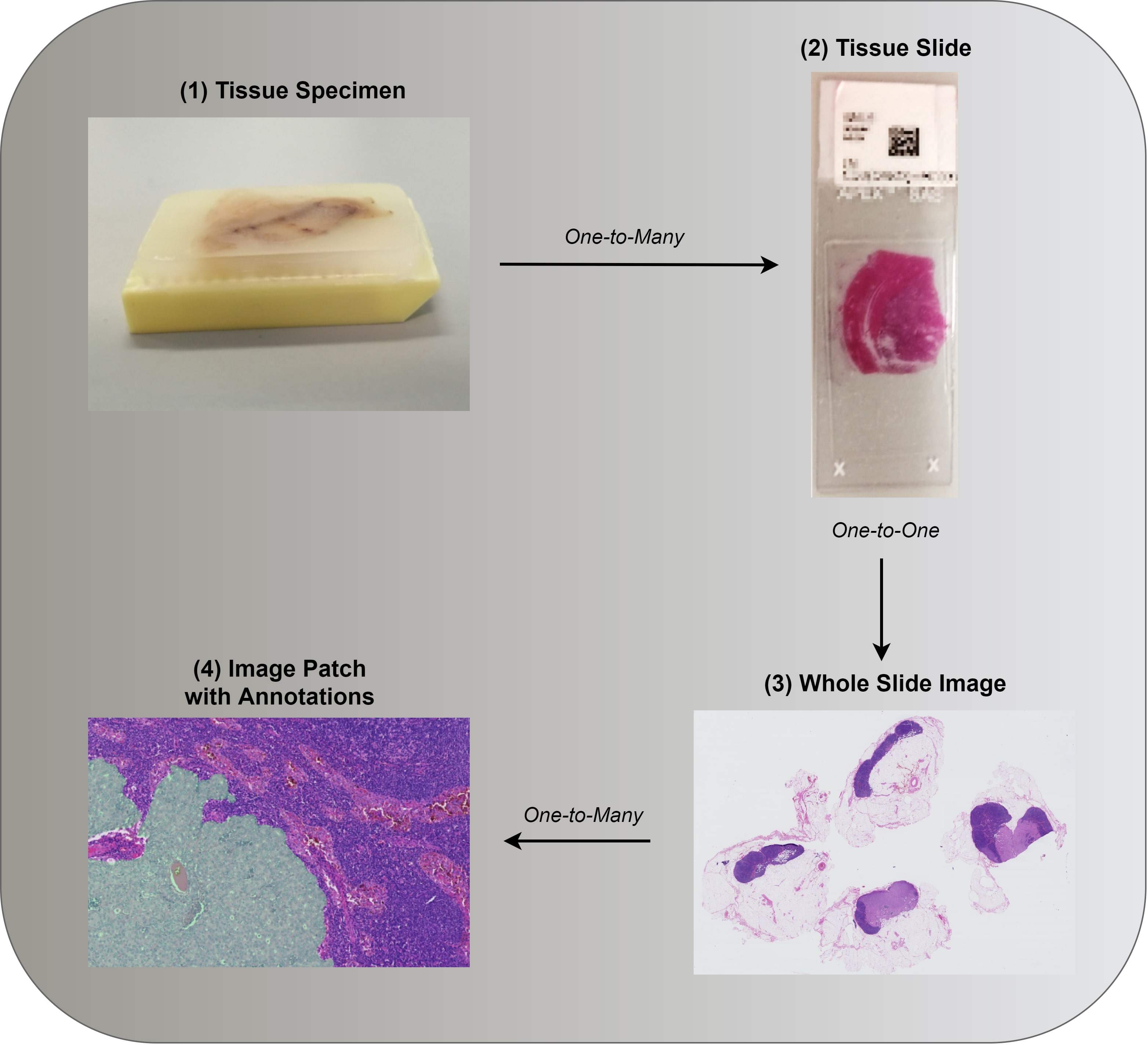}
\end{center}
\caption{(1): Tissue specimen is often investigated as a potential predictor of patient diagnosis, prognosis, or other patient level information. (2 - 3): Both in clinical practice and research, in the interest of time, a single tissue slide, or its digital counterpart, is often assessed. Annotations associated with a single tissue section can be provided such as whether a malignancy is present. (4):  Consequent to the gigapixel size of WSIs, image analysis requires further image reduction. Patches are often extracted based on annotations, if available, or otherwise (see Section~\ref{sec:slidelevel}). Images (3) and (4) were taken from the public data set of Camelyon17~\cite{BandGeesMansDijk+2019}.
}\label{fig:2}
\end{figure}

Most successful approaches to training deep learning models on WSIs do not use the whole image as input and instead extract and use only a small number of patches~\cite{JanoMada2016,EhteVetaJoha+2017,BandGeesMansDijk+2019,AresArauKwokChen+2018}. Image patches are usually square regions with dimensions ranging from $32\times32$ pixels up to $10000\times10000$ pixels with the majority of approaches using image patches of around $256\times256$ pixels~\cite{JanoMada2016,AresArauKwokChen+2018,ChanJungWooLee+2019}. This approach to reducing the high dimensionality of WSIs can be seen as human guided feature selection. The way patches are selected constitutes one of the key areas of research for WSI analysis. Existing approaches can be grouped based on whether they employ annotations and at which level (see Figure~\ref{fig:2}). 

\subsubsection{Patch level annotation}
Patch level annotations enable strong supervision since all of the extracted training patches have class labels. Typically, patch based annotations are derived from pixel level annotations which requires experts to annotate all pixels. For instance, given a WSI which contains cancerous tissue, a pathologist would need to localize and annotate all cancerous cells.

A simple approach to patch based learning would make use of all tiled (i.e.\ non-overlapping) patches. Nevertheless, this simplicity comes at the cost of excessive computational and memory overhead, along with a number of other issues, such as imbalanced classes and slow training. Randomly sampling patches may lead to an even higher class imbalance considering how in most cases a patch is much smaller than the original WSI. It is therefore imperative that sampling is guided. 

One way to guide the sampling procedure is with the use of patch level annotations. For example, on breast cancer metastasis detection, in multiple papers, patches from normal and tumour regions were extracted based on pixel level labels that were provided by pathologists ~\cite{LiuGadeNoroDahl+2017,WangKhosGargIrsh+2016,LiPing2018,KongWangLiSong+2017,ZanjZingdeWi2018,EhteVetaJoha+2017,BandGeesMansDijk+2019}. Others were able to detect, segment, and classify different types of cell nuclei, colon glands and other large organs, as well as to classify and localize a variety of diseases~\cite{LitjKooiBejnSeti+2017,CoudOcamSakeNaru+2018,ChanJungWooLee+2019}. Most successful approaches also employ hard negative mining, an iterative process whereby false positives are added to the training data set for further training~\cite{WangKhosGargIrsh+2016,LiPing2018,EhteVetaJoha+2017,BandGeesMansDijk+2019}. Identification of false positives is possible in such cases due to the availability of patch level annotations.

\subsubsection{Slide level annotation}
\label{sec:slidelevel}
Due to practical limitations, in most cases ground truth labelling is done on the level of WSIs as opposed to individual patches. Despite this lower granularity of labelling, a number of deep learning based approaches have demonstrated highly promising results. Techniques vary and often take on the form of multiple instance learning, unsupervised learning, reinforcement learning, and transfer learning, or a combination of thereof. Intuitively, the goal is usually to identify patches that can collectively or independently predict the whole slide label. 

Preprocessing based on image filters can be employed to reduce the number of patches that need to be analysed. Multiple studies also employ the Otsu, hysteresis, or other types of threshold as an automatic way of identifying tissue within the WSI. Other operations such as contrast normalisation, morphological operations, and a problem specific patch scoring system can also be employed to reduce further the number of candidate patches and even enable automatic patch localization. However, verifying that indeed each patch has the same label as the slide often requires domain-specific expertise and even the process of coming up with the best image filters requires at the very least some human intuition.

In order to avoid potential human bias, most approaches employ unsupervised or multi-instance learning, or a combination of both. Tellez et al.\ \cite{TellLitjLaakCiom2018} examined different methods of unsupervised representation learning for compressing image patches into a lower dimensional latent space. It was then possible to train CNNs directly on the compressed WSIs~\cite{TellLitjLaakCiom2018}. Others reduced the dimensionality of patches using traditional dimensionality reduction techniques, such as principal component analysis, as well as CNNs pretrained on ImageNet~\cite{ZhuYaoZhuHuan2017,YueDimiCaieHarr+2019}.  Both Zhu et al.\ and Yue at al.\ \cite{ZhuYaoZhuHuan2017,YueDimiCaieHarr+2019} subsequently used $k$-means clustering and found the most discriminative clusters of patches by training CNNs in a weakly supervised manner~\cite{Zhou2017}. However, the multi-stage structure of the aforementioned techniques does not allow processes that come first, i.e.\ patch compression or patch localization, to improve following the improvement of later processes, i.e.\ visual understanding.

Several other ideas enable persistent improvement of patch localization and visual understanding either by iteratively revising each process or by learning both in an end-to-end fashion. Hou et al.~\cite{HouSamaKurcGao+2016} proposed an expectation maximisation algorithm that enables increasingly more discriminative patches to be selected while at each iteration the CNN was trained further for 2 epochs. A patch was considered more discriminative if, when given as an input to the CNN, the prediction was closer to the slide level label~\cite{HouSamaKurcGao+2016}.  Combalia et al.\ \cite{combVila2018} instead of an expectation maximisation algorithm, employed Monte Carlo sampling. One of the most promising emerging directions aims at incorporating the process of selecting patches within the optimisation of visual understanding~\cite{CourTramSansWain2018,IlseTomcWell2018,QaisRajp2019,AichGhas2018,MomeThibGeva2018,TomiAbdoWeiRen+2018}.  Courtiol et al.\ \cite{CourTramSansWain2018} modified a pretrained ResNet-50 by adding a $1\times1$ convolutional layer after the convolutional layers to get patch level predictions. A MinMax layer was added on top followed by fully connected layers to predict the slide level label~\cite{CourTramSansWain2018}. The MinMax layer is a type of attention mechanism which gives the capability of selective training on the most discriminative patches of both classes. 

Instead of extracting features from all or most patches before selecting a few to learn on, recent work has employed attention models~\cite{AichGhas2018,QaisRajp2019,MomeThibGeva2018,TomiAbdoWeiRen+2018}. Intuitively, an attention model is initially as good as random guessing at patch selection but progressively chooses more discriminative patches that contribute to better model performance. For example, Qaiser et al.\ \cite{QaisRajp2019} used reinforcement learning to train a model in selecting patches at $20\times$ and $10\times$ magnification levels based on a low resolution image at $2.5\times$ magnification level. Using supervised learning, BenTaieb et al.\ \cite{AichGhas2018} employed a recurrent visual attention network that processes non-overlapping patches of $5,000\times5,000$ pixels at $20\times$ magnification level and sequentially identifies and analyses different regions within those patches.

\subsubsection{Patient level annotation}
Usually, multiple WSIs can be acquired for each patient since the initial tissue occupies a 3D space and therefore multiple cuts can be made. In this case, the available ground truth can be specific to the patient, but not to each individual WSI~\cite{ZhuYaoZhuHuan2017,YueDimiCaieHarr+2019}. This is typically addressed based on the same operations that are used when aggregating patch level predictions to slide level predictions~\cite{ZhuYaoZhuHuan2017,YueDimiCaieHarr+2019}.

\subsubsection{Aggregating to a higher level}
In many cases, training takes place at a lower level, e.g.\ patch level, but the end goal resides at a higher level, e.g.\ slide level. For example, in the case of cancer diagnosis, a CNN may be trained to identify the presence of cancerous cells within a patch. However, some type of aggregation is needed in order to infer whether a WSI contains cancerous cells. This may take the form of a maximum or average operation over some or all patch predictions. In other cases, traditional machine learning models, or recurrent neural networks may be employed and trained using features extracted by a CNN and the ground truth that is available at a higher level.
\subsection{Beyond patch extraction}
A primary limitation of patch based analysis emerges as a consequence of analysing a large input image by means of independent analysis of smaller regions. In particular, such approaches are inherently unable to capture information distributed over scales greater than the patch size. For example, although cell characteristics can be extracted from individual patches, higher level structural information, such as the shape or extend of a tumour, can only be captured when analysing larger regions. Explicitly modelling spatial correlations between patches has been proposed as a potential solution~\cite{KongWangLiSong+2017,LiPing2018,ZanjZingdeWi2018}. However, this idea has only been tested with a small number of neighbourhoods and requires patch level annotations. A different approach involves patch extraction from multiple magnification levels~\cite{HouSamaKurcGao+2016,LiuGadeNoroDahl+2017,CampSilvFuch2018}. Others, such as the attention models described above, consider global context, that is a low resolution image of the WSI, both when choosing regions to attend to and when predicting the slide level label~\cite{QaisRajp2019,AichGhas2018}. Finally, recent work attempts to ameliorate some of the aforementioned problems associated with patch based analysis by using much larger patch sizes~\cite{AichGhas2018,MomeThibGeva2018}.

\section{Discussion}
The aim of computer vision is to create algorithmic solutions capable of visual understanding. Applications can range from object identification and detection to image captioning and scene decomposition. In the past decade most areas of computer vision have seen remarkable progress, much of it effected by advances in neural network based learning algorithms~\cite{GuoLiuOerlLao+2016,VoulDoulDoulProt2018}. The success of these methodologies, part of the now established field of deep learning, can be attributed to a number of reasons, with the transformation of the feature extraction stage often described as the leading factor. 

In the previous decade most approaches focused on finding ways to explicitly extract features from images for models subsequently to employ~\cite{Lowe2004,DalaTrig2005}. Therefore, feature extraction and model development were two distinct, independent stages that were performed sequentially, and where the former was based on human intuition of what constitutes a good feature. Automating this process through the use of convolutional neural networks (CNNs) has been shown to result in more discriminative features tailored for the problem at hand~\cite{EhteVetaJoha+2017,AresArauKwokChen+2018,QuasMukhReddMunu+2017,VetaHengStatBejn+2019}. This is one of the reasons behind the success of deep learning, and more broadly, neural network based learning, as feature extraction became a learning process, fundamentally intertwined with the learning of model parameters. Some of the other key factors which contributed to the successes of deep learning include advancements in hardware and software, as well as the increase in data availability.

The analysis of multi-gigabyte images is a new challenge for deep learning that has only appeared along the emergence of digital pathology and whole slide imaging. Building deep learning models capable of understanding WSIs presents novel challenges to the field. When patch level labels are available, patch sampling coupled with hard negative mining can train deep learning models that in many cases match and even surpass the accuracy of pathologists~\cite{QuasMukhReddMunu+2017}. For many medical data sets with patch level annotations, deep learning models seem to excel, and with the introduction of competitions, such as Camelyon16 and Camelyon17~\cite{EhteVetaJoha+2017,BandGeesMansDijk+2019}, this type of deep learning has repeatedly demonstrated its success in performance and interpretability~\cite{JanoMada2016}. Therefore, patch based learning from gigapixel images and patch level annotations seems to be the closest to clinical employment. However, in many cases, only labelling with lower granularity can be attained either because it is very laborious and expensive, or simply because it is infeasible. In addition, patch level supervision may be limiting the potential of deep learning models as the models can only be as good as the annotations provided.

To work with slide or patient level labels, current approaches focus on the \textit{where} problem, or in other words, on approximating the spatial distribution of the signal~\cite{TellLitjLaakCiom2018}. Out of the work we reviewed, only Tellez et al.~\cite{TellLitjLaakCiom2018} has instead simplified the \textit{what} problem, i.e.\ visual understanding, to the point where the \textit{where} problem becomes trivial. It would be interesting to see the efficacy of the work by Tellez et al. on harder problems, such as prognosis estimation, and with other low dimensional latent mappings. On the \textit{where} problem, there are generally two approaches. The first uses a type of meta-learning, where in order to optimise the \textit{where} problem, the \textit{what} problem has to first be optimised. The second approach attempts to optimise both \textit{what} and \textit{where} problems simultaneously in an end-to-end setting. This is done by either forwarding a set of patches through a CNN and attending on a few or by localizing and attending to a single patch at each time step. 

Deep learning is already demonstrating its potential across a wide range of medical problems associated with digital pathology. However, the need for detailed annotations limits the applicability of strongly supervised techniques. Other techniques from weakly supervised, unsupervised, reinforcement, and transfer learning are employed to counter the need for detailed annotations while dealing with massive, highly heterogeneous images and small data sets. This emerging direction away from strong supervision opens new opportunities in WSI analysis, such as addressing problems for which the ground truth is only known at a higher than patch level, e.g.\ patient survivability and recurrence prediction.

\bibliography{./wsi_dl}
\end{document}